\documentclass[journal,twoside,web]{ieeecolor}
\usepackage{generic}
\usepackage{cite}
\usepackage{amsmath,amssymb,amsfonts}
\usepackage{algorithmic}
\usepackage{graphicx}
\usepackage{textcomp}
\usepackage{multirow}
\usepackage[switch]{lineno}
\def\BibTeX{{\rm B\kern-.05em{\sc i\kern-.025em b}\kern-.08em
    T\kern-.1667em\lower.7ex\hbox{E}\kern-.125emX}}
\begin{document}
\title{Affinity Feature Strengthening for Accurate, Complete and Robust Vessel Segmentation}
\author{Tianyi Shi, Xiaohuan Ding, Wei Zhou, Feng Pan, Zengqiang Yan, Xiang Bai and Xin Yang, \IEEEmembership{Member, IEEE}
\thanks{Tianyi Shi, Xiaohuan Ding, Zengqiang Yan and Xin Yang are with the School of Electronic Information and Communications, Huazhong University of Science and Technology, Wuhan, China. Xiang Bai is with the School of Artificial Intelligence and Automation, Huazhong University of Science and Technology, Wuhan, China. Xin Yang is the corresponding author. (e-mail:xinyang2014@hust.edu.cn). }
\thanks{Wei Zhou is with Cloud BU, Huawei Technologies Company Limited, Shenzhen, China.}
\thanks{Feng Pan is with the Department of Radiology, Union Hospital, Tongji Medical College, Huazhong University of Science and Technology, and with Hubei Province Key Laboratory of Molecular Imaging, Wuhan, China.}
\thanks{This work is supported by the National Natural Science Foundation of China (62061160490, 62122029, U20B2064).}
}

\maketitle

\begin{abstract}
Vessel segmentation is crucial in many medical image applications, such as detecting coronary stenoses, retinal vessel diseases and brain aneurysms. However, achieving high pixel-wise accuracy, complete topology structure and robustness to various contrast variations are critical and challenging, and most existing methods focus only on achieving one or two of these aspects. In this paper, we present a novel approach, the affinity feature strengthening network (AFN), which jointly models geometry and refines pixel-wise segmentation features using a contrast-insensitive, multiscale affinity approach. Specifically, we compute a multiscale affinity field for each pixel, capturing its semantic relationships with neighboring pixels in the predicted mask image. This field represents the local geometry of vessel segments of different sizes, allowing us to learn spatial- and scale-aware adaptive weights to strengthen vessel features. We evaluate our AFN on four different types of vascular datasets: X-ray angiography coronary vessel dataset (XCAD), portal vein dataset (PV), digital subtraction angiography cerebrovascular vessel dataset (DSA) and retinal vessel dataset (DRIVE). Extensive experimental results demonstrate that our AFN outperforms the state-of-the-art methods in terms of both higher accuracy and topological metrics, while also being more robust to various contrast changes. The source code of this work is available at https://github.com/TY-Shi/AFN.
\end{abstract}


\section{Introduction}
\label{sec:introduction}

\begin{figure}[h]
\centerline{\includegraphics[width=0.95 \columnwidth]{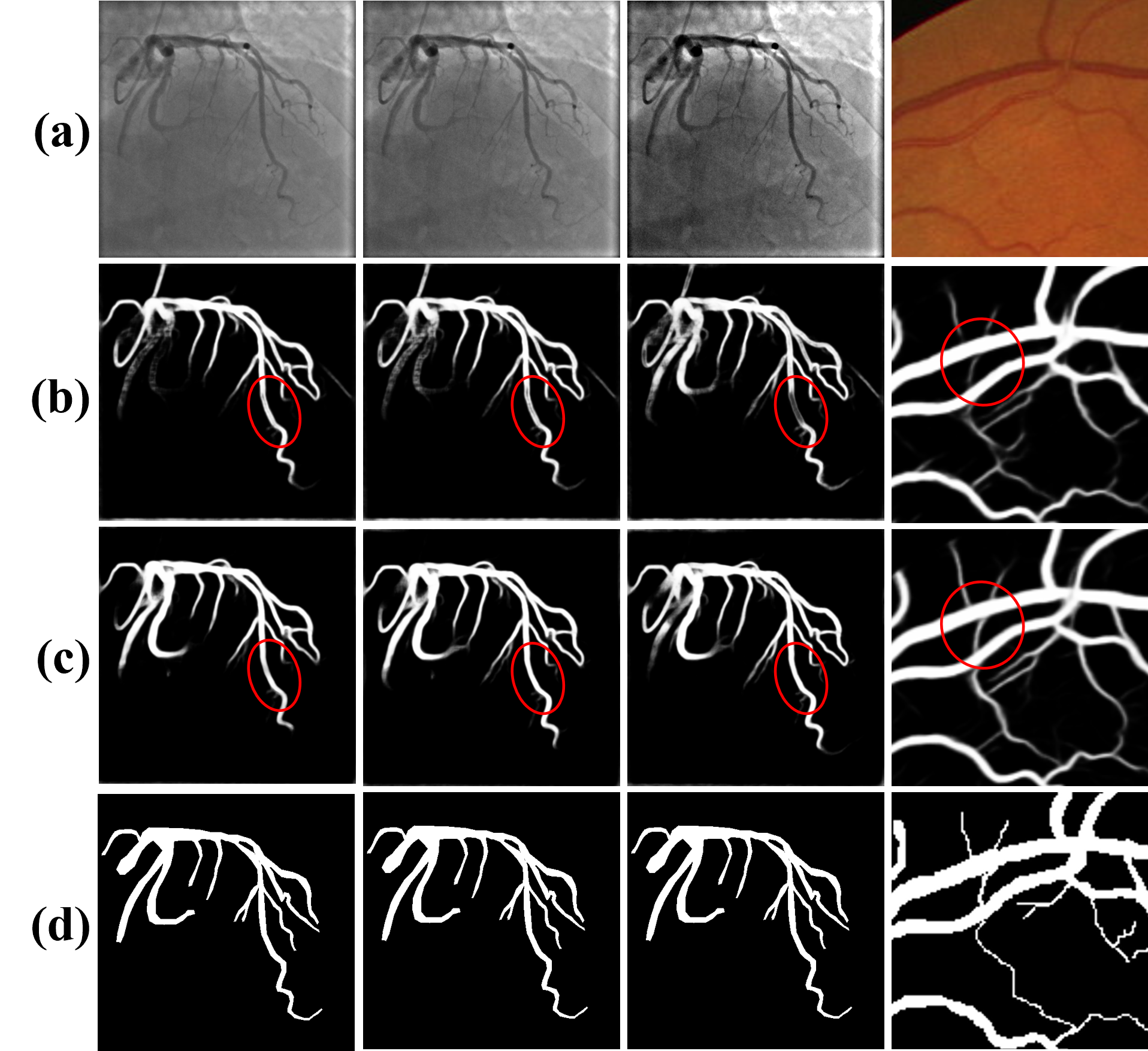}}
\caption{(a) original images (columns 1 to 3 are XCAD images with varying contrast, and the last column is retinal vessel images). (b)-(d) segmentation result of the state-of-the-art method\cite{cheng2021joint}, our method and ground truth respectively. Our method produces more accurate pixel-level results, complete topology structure and is insensitive to contrast variations.}
\label{problem}
\end{figure}

\IEEEPARstart{V}{essel} segmentation, which aims to obtain both accurate delineations of vessel boundaries (i.e., pixel-wise accurate) and complete vascular structures (i.e., topology-preserving), is critical for many medical applications. For instance, accurately delineating vessel boundaries from angiography images can help detect coronary stenoses\cite{zhang2020direct} and cerebral aneurysms\cite{su2021autotici,zeng2019automatic}, and hemodynamic analysis\cite{tahir2020anatomical}, and Alzheimer's disease diagnosis\cite{haft2019deep}, and generating computational anatomical\cite{damseh2020laplacian}, and assist interventional operation planning\cite{van2022automatic}. Obtaining a complete topology of vessels from retinal images can facilitate early detection of several retinal diseases, such as diabetic retinopathy\cite{archer1999diabetic} and age-related macular degeneration\cite{li2021applications}. Meanwhile, given large variations present in medical images due to different acquisition equipment and imaging procedures, we desire our method can be invariant to large contrast changes to ensure consistently high performance for a wide range of medical images.

Despite substantial research in the literature, concurrently achieving pixel-wise accurate, topology-preserving and contrast-insensitive vessel segmentation remains highly challenging due to highly complex vascular structures, blurred boundaries in particular for thin vessels and large quality variations in practical medical images, as shown in Fig.~\ref{problem}. There are several studies\cite{hu2019topology,mosinska2018beyond,hu2020topology,shit2021cldice,tan2022retinal, cheng2021joint} aiming at preserving the complete topology of curvilinear structures. For instance, Topoloss\cite{hu2019topology} designs an explicit topology loss based on holes and connected components. However, Topoloss optimizes only limited pixels for topology and thus is sensitive to complex backgrounds. Mosinska et al.\cite{mosinska2018beyond} assume features produced by pre-trained VGG naturally encode topology information. Such an assumption may not hold in practice and thus greatly limits performance. Tan et al.\cite{tan2022retinal} propose a skeleton fitting module to capture and preserve the morphology of vessels. However, they only use skeletons as additional supervision constraints which may fail to derive explicit geometric structure representations for vessels. JTFN\cite{cheng2021joint} designs a two-stream network to jointly learn boundary detection and semantic segmentation. The learned boundary features are treated as geometric constraints to refine the vessel features of the semantic segmentation stream. However, geometric structures such as boundaries and skeletons lack clear semantic information, which limits the performance improvement for obtaining complete topology.

Most existing vessel segmentation methods\cite{ma2020rose,mou2019dense,li2022dual,gu2019net, li2020accurate,liu2022full,li2020accurate,damseh2018automatic} adopt convolutional neural networks (CNNs) which utilize pixel-wise loss functions\cite{ronneberger2015u} to learn multiscale features for vessels of different sizes. As the number of pixels for thin vessels is typically much smaller than that of thick ones, learning a CNN via pixel-wise losses inevitably biases to accurately segmenting thick vessels yet ignoring thin vessels, degrading the completeness of the vascular topology. To address this issue, Yan et al.\cite{yan2018three} propose a three-stage deep learning model to learn distinctive features for thick and thin vessels separately at different stages. The segmentation results of thick and thin vessels are then merged at the last stage. Several other researchers\cite{shit2021cldice,yan2018joint} propose to train a CNN via both pixel-wise and topology-wise losses. For instance, Yan et al.\cite{yan2018joint} train a CNN using both the segment-level and pixel-wise losses to balance the segmentation performance of thick and thin vessels. Shit et al.\cite{shit2021cldice} introduce the centerline dice loss based on center-line matching to improve topological completeness. However, the above methods cannot concurrently handle topology for vessels of different scales. Moreover, vessel features they learned purely rely on image intensities and in turn could be sensitive to contrast changes.

To improve the robustness of segmentation models to illumination and/or contrast changes, which widely exist in medical images, existing methods exploit data augmentation\cite{sun2021robust} or some dedicated approaches. For instance, in\cite{li2017novel} the authors propose a convex-regional-based gradient model which computes the differences between the intensity summation in the symmetrical pixel intervals. Their method can reduce the sensitivity to contrast by constructing relative differences between pixel intensities. Shi et al.\cite{shi2022local} design a pre-processing approach that converts image intensities to relative pixel difference representations to counteract contrast changes. To sum up, to confront illumination and contrast changes, it is inspiring to study how to utilize relative information among pixels and reduce the reliance on absolute intensities to model topological structures and refine vessel features.

In this paper, we propose an affinity feature strengthening network (AFN), as shown in Fig.~\ref{pipeline}, towards a vessel segmentation model that can concurrently achieve high pixel-level accuracy, complete topology and robustness to contrast changes. The core of our AFN is a novel supervised multi-scale affinity feature strengthening module (SMAFS) which learns a set of affinity fields from the predicted segmentation mask to capture the semantic relations among pixels.  Such semantic relationships can well reflect the geometric structure (as shown in Fig.~\ref{Affinity_D}) and in turn implicitly model the topology of a vascular tree. The affinity fields are then used to strengthen and refine segmentation-related image features for vessels with different sizes. As the ground-truth affinity field is only available for the full-size image, we learn the full-size affinity field explicitly under the guidance of an affinity cosine distance (ACD) loss between the predicted multi-scale affinity fields and its ground truth. For affinity fields of intermediate layers, we learn the affinity implicitly via the unsupervised affinity feature strengthening module (UAFS). To encode pixel relations for vessels of different sizes, we calculate multi-scale affinity fields and establish the affinity relationship at multiple layers. That is, we compute affinity fields of pixels within different-scales and learn spatial- and scale-aware adaptive weights for affinity fields to better enhance vessel features at different scales. As the affinity considers only the relative classification labels of neighboring pixels, it captures the intrinsic geometric and contextual information of an object and is robust to contrast variations.

We evaluate our method on two public datasets, i.e., the X-ray angiography coronary vessel dataset (XACD)\cite{ma2021self}, the retinal vessel dataset (DRIVE)\cite{maninis2016deep}, and two in-house datasets, i.e., a digital subtraction angiography cerebrovascular vessel dataset (DSA) and a portal vein dataset (PV). Extensive experimental results demonstrate the superiority of our AFN to the state-of-the-art methods\cite{mou2021cs2,hu2019topology,cheng2021joint,ke2018adaptive}.

\section{Related Works}
\label{sec:Related Works}
\subsection{Vessel segmentation}
The goal of vessel segmentation is to obtain both accurate delineations of vessel boundaries and complete vascular structures. Towards this goal, existing methods can be roughly classified into three classes: 1) semantic segmentation which focuses on achieving high pixel-level accuracy, 2) topology preserving methods which aim to obtain complete topology and 3) hybrid methods.

Methods rely on semantic segmentation mainly adopt a U-Net structure as the backbone and focus on integrating context features\cite{mou2019dense,gu2019net,mou2019cs,mou2021cs2,ding2020high,wang2020csu} and/or feature enhancement\cite{ma2020rose,li2022dual,zhang2020befd} methods to improve their ability for capturing representative tubular structure of vessels. Specifically, \cite{mou2019dense,gu2019net,ding2020high,xu2020joint,rodrigues2020element} extract context semantic information from larger receptive fields (e.g., dilated convolution) and learn different position/region pixels' correlation and difference. \cite{mou2019dense,gu2019net} apply various dilated convolutions to enlarge fields. In \cite{ding2020high}, authors exploit a non-local-like module to obtain a global attention map that models the relationship among all pixels for vessel segmentation. CS$^{2}$Net\cite{mou2019cs,mou2021cs2} introduces spatial and channel attention to integrate local features with their global dependencies. Yan et al.\cite{yan2018three} propose a deep learning model to learn discriminative features for different size vessel regions, and utilize joint segment-level and pixel-wise losses\cite{yan2018joint} to reduce the imbalanced pixel ratio between thick and thin vessels. Feature enhancement methods utilize various kinds of geometric structural information. \cite{ma2020rose} learns centerline-level segmentation map to represent complex geometric structures for vessel feature enhancement. \cite{li2022dual} and \cite{zhang2020befd} propose to extract boundaries as structure priors to improve the representation ability of vessel features. However, these methods primarily focus on pixel-wise prediction accuracy and are intrinsically topology-agnostic. It is inspiring to study better geometric structure information to effectively model the topology of vessels for segmentation feature enhancement. 

Topology-preserving methods aim to achieve high completeness of vascular trees. Existing topology-preserving methods can be categorized into two groups: implicit\cite{mosinska2018beyond,shit2021cldice,gur2019unsupervised} and explicit\cite{hu2019topology,hu2020topology,damseh2018automatic}. Mosinska et al.\cite{mosinska2018beyond} rely on the pre-trained VGG to implicitly encode the topology of objects, resulting in limited performance improvement. Gur et al.\cite{gur2019unsupervised} integrate the active contour model and a learning-based model to get the vessel structure. The authors in\cite{shit2021cldice} propose a centerline loss which measures the similarity between the predicted vessel centerlines and the ground truth centerlines to implicitly preserve topology. In Topoloss\cite{hu2019topology}, the authors learn the local topology explicitly based on detected holes and connected components. Damseh et al.\cite{damseh2018automatic} propose a geometry contraction algorithm for segmentation result as post-processing to refine the topology structure.
In this work, we aim to model topology for feature enhancement yet improves the robustness of the topology-preserving solution to contrast variations.

In the literature, several methods\cite{cheng2021joint,tan2022retinal} focus on achieving both pixel-level and topology-level accuracy. For instance, JTFN\cite{cheng2021joint} develops a two-stream network that leverages a boundary detection branch to establish the relationship between topology and boundary connectivity and then uses the topology information to enhance segmentation-related features. Tan et al.\cite{tan2022retinal} propose a skeleton fitting module to capture the morphology of the vessels and utilizes such morphological information as additional constraints in training. Motivated by these methods, in this work we try to construct relationships between affinity fields and topology, and utilize affinity fields to enhance segmentation-related features.

\begin{figure*}[t]
	\centerline{\includegraphics[width=\linewidth]{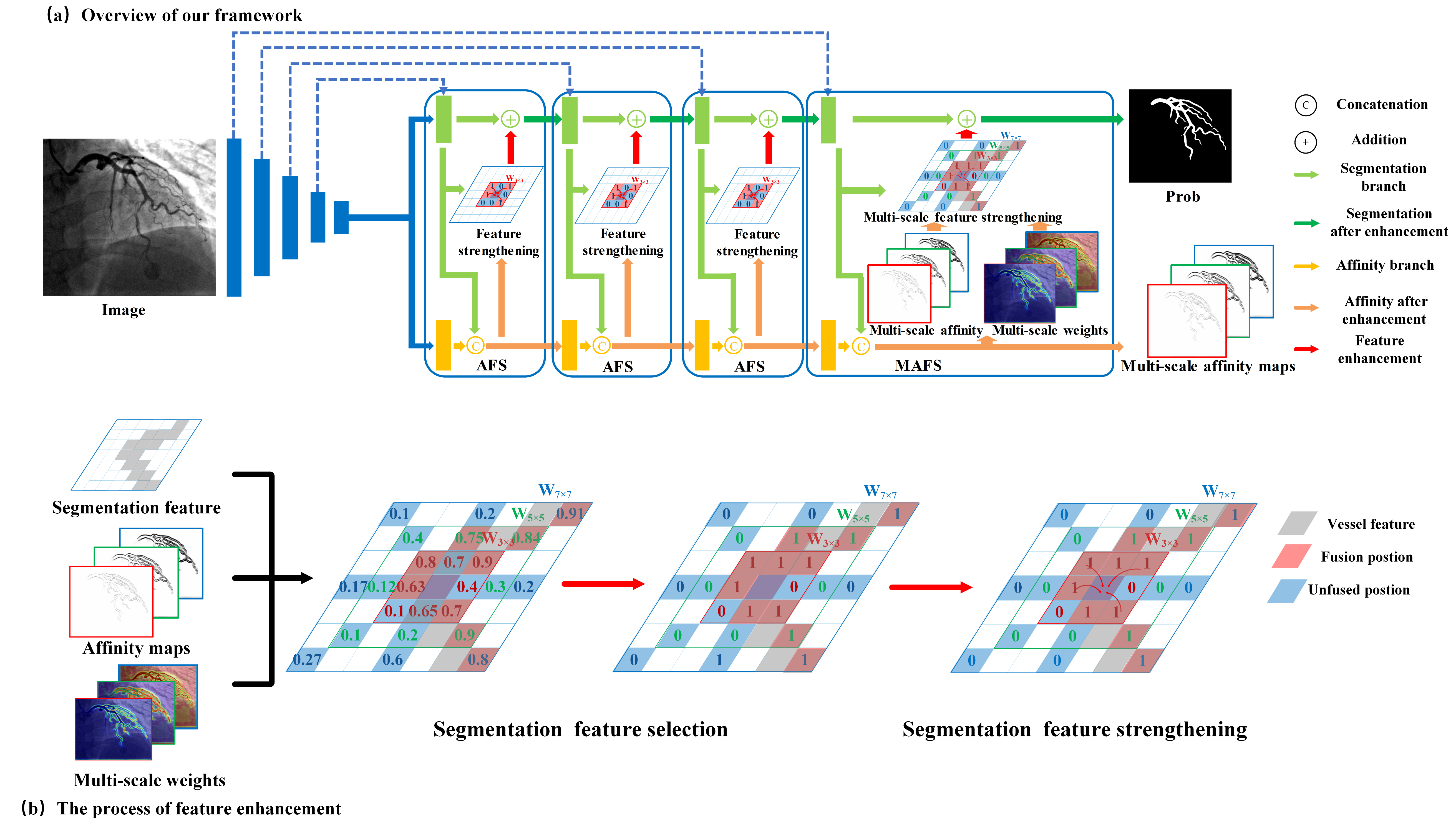}}
	\caption{The pipeline of Affinity Feature Strengthening Network (AFN). (a) The overview of AFN. (b) The process of multi-scale affinity field enhancement.}
	\label{pipeline}
\end{figure*}

\subsection{Affinity Field Feature learning}
Affinity fields, explicitly encoding the semantic relationships among neighboring pixels (i.e. whether or not a pixel is from the same category with its neighboring pixels), can implicitly represent the local topology of objects\cite{ru2022learning,zhang2021affinity} as shown in Fig.~\ref{Affinity_D}. Compared to existing methods which encode topology using skeletons\cite{tan2022retinal} and boundaries\cite{cheng2021joint,li2022dual}, the affinity fields rely on intrinsic semantic label relationship instead of the absolute pixel intensity and thus are insensitive to illumination/contrast changes. In addition, affinity fields which encode semantic guidance are more instructive than the attention methods\cite{mou2019cs,ma2020rose,zhang2020befd,mou2021cs2,wang2020hard,yuan2021multi,yan2020attention} for feature enhancement. In general semantic segmentation tasks, existing methods based on affinity field feature learning either establish the semantic relationship between pixels to refine segmentation results\cite{gao2019ssap,liu2018affinity,ahn2018learning,ru2022learning}, or introduce additional loss constraints\cite{ke2018adaptive,zhang2021adaptive}. For instance, \cite{gao2019ssap} utilizes a learned affinity pyramid to gradually refine the segmentation results from coarse to fine. AffinityNet\cite{ahn2018learning} considers pixel-wise affinity to learn comprehensive semantic information for refining initial segmentation pseudo labels. AFA\cite{ru2022learning} also learns affinity fields to refine the initial pseudo labels for segmentation with Transformers. Several other researchers\cite{ke2018adaptive,zhang2021adaptive} characterize the geometric structures by affinity field loss constraints. For example, AAF\cite{ke2018adaptive} proposes an adaptive affinity field loss function (AAF) to capture the semantic relations between neighboring pixels in the label space. Zhang et al.\cite{zhang2021adaptive} further improve AAF with different region weights to learn reliable relationships for weakly supervised semantic segmentation. Different from existing methods which use affinity fields to refine or constrain segmentation results, our AFN leverages affinity fields to enhance intermediate features at different layers in the feature space. We demonstrate that our AFN can make full use of the affinity fields to establish semantic correlation and encode geometric structure between pixel features, and in turn achieve promising performance in terms of pixel-level accuracy, topological completeness, and robustness to contrast changes.

\subsection{Contrast-insensitive feature learning}
To improve the robustness of segmentation models to contrast changes, existing methods mainly utilize data augmentation\cite{sun2021robust}, contrast-invariant handcrafted features\cite{li2017novel,kassim2019deep}, and image preprocessing approaches\cite{shi2022local}. Sun et al.\cite{shi2022local} propose two new data augmentation modules, i.e., channel-wise random gamma correction and random vessel augmentation, to improve the robustness of vessel segmentation. Li et al.\cite{li2017novel} propose a new convex-regional-based gradient feature to describe the differences between the intensity summation in the symmetrical intervals to reduce the contrast sensitivity in a near-infrared image. \cite{kassim2019deep} combines several kinds of handcrafted features (intensity, orientation and curvature, etc.) and U-Net regression prediction results to improve the segmentation robustness. LIOT\cite{shi2022local} proposes a local intensity order transformation approach to improve the generalizability. Inspire by these methods, we model the topology information without reliance on raw intensities to improve the robustness to contrast variations.

\section{Method}
\label{sec:method}
In this section, we present in detail the AFN architecture, which consists of a supervised multi-scale affinity feature strengthening (SMAFS) module and three unsupervised affinity feature strengthening (UAFS) modules, as illustrated in Fig.~\ref{pipeline}. Details of each module are discussed in the following.
\subsection{Overview}
AFN is designed in a U-Net like shape, where the encoder is defined as\cite{cheng2021joint} consisting of the first five layers of VGG16\cite{simonyan2014very} to capture five hierarchical CNN features for an input image and the decoder consisting of a four-layer two-branch structure to reconstruct pixel-wise semantic features and affinity fields respectively as shown in Fig.~\ref{pipeline}(a). Between the segmentation and affinity branches, we introduce one SMAFS module and three UAFS modules to learn the affinity relationship of neighboring pixels and then utilize the affinity information as geometric and context guidance to enhance the segmentation features. To learn different size spatial structures of a vascular tree, we propose SMAFS to adapt various vessels and backgrounds by learning multi-scale full-size affinity fields explicitly. In this way, vessel segments of different scales can be well captured by SMAFS. In addition, SMAFS also enhances the segmentation features under affinity fields guidance to encode multi-scale structural relations for better topology-level performance. As the affinity field ground truth of features is not available due to the inability to accurately obtain the affinity relationship between the features, we further propose UAFS to learn affinity fields implicitly at intermediate layers. SMAFS and UAFS both operate on the segmentation relations instead of the absolute pixel’s features to capture the intrinsic semantic relationships regardless of visual appearance variations. Therefore, SMAFS and UAFS can better preserve the geometric structure and achieve high pixel-level accuracy while being robust to contrast variations. In addition, to mitigate the spatial loss of encoders arising from pooling operations, we add skip connections between each encoding and the corresponding decoding layers. In this step, we follow\cite{cheng2021joint} to embed a gated attentive unit in each skip connection to integrate both context- and spatial-aware predictions for better feature fusion and refinement.

\subsection{Supervised Multi-scale Affinity Feature Strengthening}
Affinity fields capture the relationship between two pixels belonging to the same category, providing a pixel-centric description of semantic relations in the space. Prior work in\cite{ke2018adaptive,ru2022learning,zhang2021affinity} has demonstrated the effectiveness of affinity fields in characterizing semantic label correlations, which can be integrated into learning as geometric and context guidance.

\begin{figure}[tbp]
	\centering{\includegraphics[width=0.5 \textwidth]{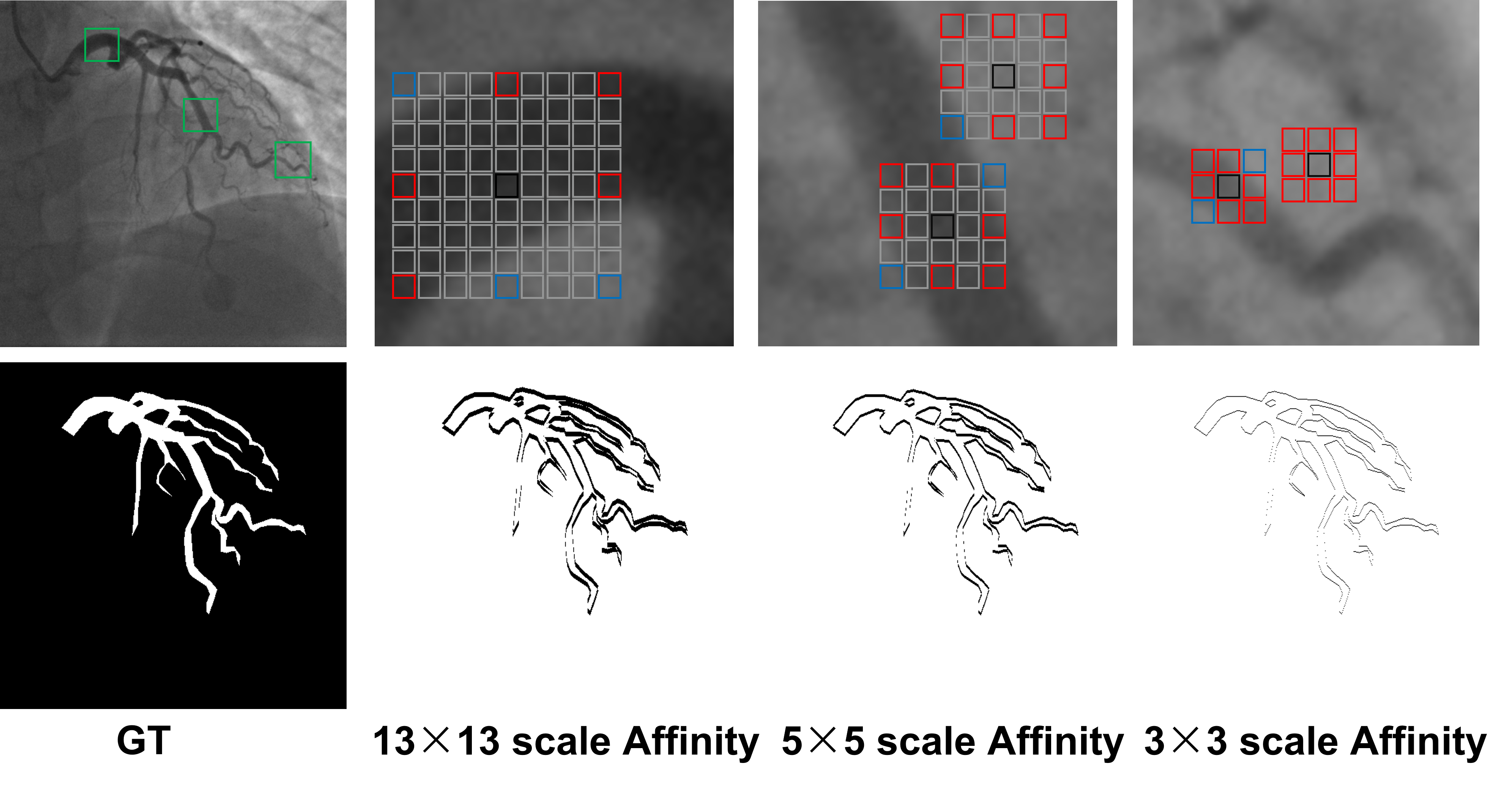}}
	\caption{Different scale affinity fields for vessels of different sizes. The green rectangular blocks indicate different local zoom regions. The red grid indicates the pixel is the same category with the central pixel, and the blue grid indicates the pixel is the different category with the central pixel. The second row shows the ground truth, 13×13, 5×5, and 3×3 scale affinity fields in the right direction respectively (These three scales are just for easy affinity field visualization-the actual scales of affinity fields affinity applied to XCAD is 3×3, 9×9, 15×15).} 
	\label{Affinity_D}
\end{figure}

However, a single-scale affinity field which considers only semantic relationships among pixels within a single fixed range cannot fit all vessel structures. To handle vessels of different sizes (e.g., thick vessels and thin vessels), we calculate multi-scale affinity fields to strengthen the segmentation features for different-sized vessels and backgrounds. Specifically, multi-scale affinity fields are defined as follows. Given a pixel $x$ in the semantic label image $G_{s} \in \mathbb{R}^{H \times W}$, $\mathcal{N}(x)$ is the set of the eight neighboring locations of $x$. Let $x_{l}$ denote one of the eight neighboring pixels in $\mathcal{N}(x)$, i.e., at the left, right, top, bottom, left top, left bottom, right top and right bottom location $l$ of the pixel $x$ as shown the by red and blue pixels in Fig.~\ref{Affinity_D}. The red grids indicate pixels that belong to the same category as the central pixel, while the blue grids represent pixels that belong to a different category.
From Fig.~\ref{Affinity_D}, we could observe that a small-scale affinity field cannot cover the thick vessels, and the large-scale affinity field cannot well represent the adjacent semantic relationship of thin vessels well. Comparatively, jointly utilizing multi-scale affinity fields can represent pixel-centric descriptions of semantic relations at different scales effectively. Thus, we set three different affinity field sizes, including 3$\times$3, 9$\times$9, 15$\times$15 with the corresponding adaptive weights for each scale affinity field per pixel, for different object sizes as shown in Fig.~\ref{pipeline}(a). 
We denote the ground-true affinity fields $G_{A}(x)=\left(g_x^1, \cdots, g_x^N\right)$ as the corresponding $N$ relative location affinity vectors of $x$ as shown in Fig.~\ref{Affinity_D}.
If pixels $x$ and $x_{l}$ belong to the same category, we define the direction of $l$ affinity $g_x^l$ as 1. Then, the affinity $g_x^l$ for each relative location $l$ of $x$ is given by Eq.~\eqref{eq:affinity_define}

\begin{footnotesize}
\begin{equation}
    g_x^l \, = \, \begin{cases}
    1& \text{$G_{s}(x)$=$G_{s}(x_{l})$}.\\
    0& \text{$otherwise$}.
    \end{cases}
    \label{eq:affinity_define}
\end{equation}
\end{footnotesize}%
For explicit affinity field learning, we utilize a binary cross-entropy loss (BCE) and propose an affinity field cosine distance (ACD) loss to produce a clean affinity field, defined as:

\begin{footnotesize}
\begin{equation}
\mathcal{L}_{A C D}\left(G_{A}\right)=1-\frac{1}{\left|G_{A}\right|} \sum_{x \in G_{A}} \frac{\mathbf{Y}_{A}(x) \cdot \mathbf{G}_{A}(x)}{\left\|\mathbf{Y}_{A}(x)\right\|\left\|\mathbf{G}_{A}(x)\right\|}
\label{eq:LACD}
\end{equation}
\end{footnotesize}%

\begin{footnotesize}
\begin{equation}
\begin{aligned}
\mathcal{L}_{B C E}(G_{s})=&-\frac{1}{|G_{s}|} \sum_{x \in G_{s}}\left[\mathbf{G}_{s}(x) \cdot \log \left(\mathbf{Y}_{s}(x)\right)\right]\\
&+\left(1-\mathbf{G}_{s}(x)\right) \cdot \log \left(1-\mathbf{Y}_{s}(x)\right)
\label{eq:BCE_seg}
\end{aligned}
\end{equation}
\end{footnotesize}%
\begin{footnotesize}
\begin{equation}
\begin{aligned}
\mathcal{L}_{B C E}(G_{A})=&-\frac{1}{|G_{A}|} \sum_{x \in G_{A}}\left[\mathbf{G}_{A}(x) \cdot \log \left(\mathbf{Y}_{A}(x)\right)\right]\\
&+\left(1-\mathbf{G}_{A}(x)\right) \cdot \log \left(1-\mathbf{Y}_{A}(x)\right)
\label{eq:BCE_affinty}
\end{aligned}
\end{equation}
\end{footnotesize}%
\begin{footnotesize}
\begin{equation}
\mathcal{L}_{t}=\mathcal{L}_{B C E}\left(\mathbf{Y}_{s}, \mathbf{G}_{s}\right)+\mathcal{L}_{B C E}\left(\mathbf{Y}_{A}, \mathbf{G}_{A}\right)+\lambda_{b} \mathcal{L}_{A C D}\left(\mathbf{Y}_{A}, \mathbf{G}_{A}\right)
\label{eq:TotalLoss}
\end{equation}
\end{footnotesize}%
where $Y_{s}(x)$ and $G_{s}(x)$ are the predicted segmentation map and the segmentation ground truth of $x$ respectively. And we denote the predicted affinity fields $Y_{A}(x)=\left(y_x^1, \cdots, y_x^N\right)$ as the corresponding $N$ relative location predicted affinity vectors of $x$. 
$|*|$ is the size measure, $||*||$ is the magnitude of the vector, and $\lambda_{b}$ is a balancing hyper-parameter. Minimizing $\mathcal{L}_{ACD}$ encourages the model to extract better multi-scale semantic relations for different-size vessels and backgrounds.
% where $y_{A}(x)$ and $y_{s}(x)$ are the predicted affinity fields and the segmentation map of $x$, \tianyi{$G_{A}(x)$ is the ground-true affinity fields combined with eight $A(x_{l})$ at different relative locations $l$ of central pixel $x$ as shown in Fig.~\ref{Affinity_D}}, $G_{s}(x)$ is the segmentation ground truth of $x$, \tianyi{$y_{A}(x)$ is the predicted affinity fields combined with eight $y_{A}(x_{l})$ at different relative locations $l$ of central pixel $x$}, $|*|$ is the size measure, $||*||$ is the magnitude of the vector, and $\lambda_{b}$ is a balancing hyper-parameter. Minimizing $L_{ACD}$ encourages the model to extract better multi-scale semantic relations for different-size vessels and backgrounds.

For feature enhancement as shown in Fig.~\ref{pipeline}(b), we first calculate the mean predicted affinity field $\mu$ of the corresponding $N$ relative location predicted affinity vectors $\left(y_x^1, \cdots, y_x^N\right)$ of $x$, as a reference for segmentation feature selection in Fig.~\ref{pipeline}. The mean affinity field represents the overall state of the semantic relations of its neighbors, defined as

\begin{footnotesize}
\begin{equation}
% \mathbf{y}_{A_{\text {mean }}}(x)=\frac{1}{N} \sum_{x_{l} \in \mathcal{N}(x)} \mathbf{y}_{A}(x_{l})
\mathbf{\mu}(x) = \frac{1}{N} \sum_{l \in \mathcal{N}(x)} {y_x^l}
\label{eq:Amean}
\end{equation}
\end{footnotesize}%
where $N$ is the total number of neighbors of $x$. Under the guidance of the mean affinity field $\mu$, any relative position $l$ affinity $y_x^l$ greater than $\mu$ would be more likely to have similar semantic information with $x$. Then, feature strengthening is accomplished by grouping/selecting the features that are more likely to be the same category. To this end, we denote the similar category location affinity field $D_{A}(x)=\left(d_x^1, \cdots, d_x^N\right)$ as the corresponding $N$ relative location affinity vectors of $x$ by comparing each of its eight-channel location affinity fields $y_x^l$ and the mean affinity field $\mu$ to select the positions with similar category as $x$, defined as
% \begin{footnotesize}
% \begin{equation}
% \mathbf{y}_{A_{l}^{\prime}}(x)=\mathbf{y}_{A_{l}}(x)-\mathbf{y}_{A_{\text {mean }}}(x)
% \label{eq:Amean}
% \end{equation}
% \end{footnotesize}

\begin{footnotesize}
\begin{equation}
% \mathbf{y}_{A_{s l}}(x)= \begin{cases}1 & \mathbf{y}_{A_{l}}(x)-\mathbf{y}_{A_{\text {mean }}}(x) \geq 0 \\ 0 & \text { otherwise }\end{cases}
d_x^l= \begin{cases}1 & y_x^l-\mu(x) \geq 0 \\ 0 & \text { otherwise }\end{cases}
\label{eq:Hard}
\end{equation}
\end{footnotesize}%
Given the selected positions $l$, the segmentation features $f_{seg}(x)$ of $x$ are enhanced according to

\begin{figure}[tbp]
	\centering{\includegraphics[width=0.5 \textwidth]{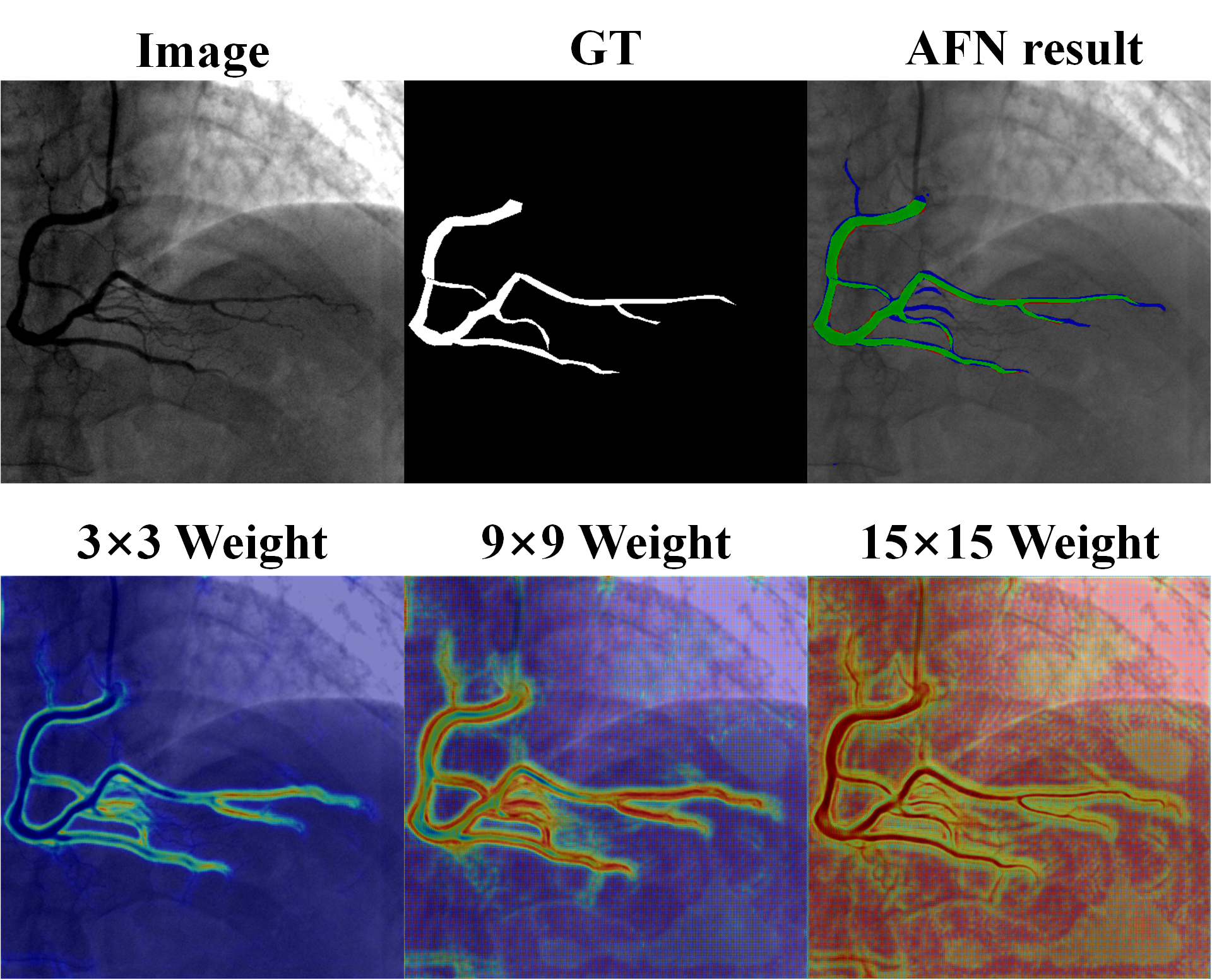}}
	\caption{Different scale weights $W_{M}$ for feature strengthening learned by SMAFS on the XCAD dataset.}
	\label{weight}
\end{figure}

\begin{footnotesize}
\begin{equation}
% f_{\mathrm{s}}(x)=\sum_{M \in \text { Sizes }} \sum_{l \in N(x)} W_{M}(x) \cdot y_{A_{s l}}(x) \cdot f_{\text {seg }}\left(x_{l}\right)+f_{\text {seg }}(x)
f_{\mathrm{s}}(x)=\sum_{M \in  S} \sum_{l \in N(x)} W_{M}(x) \cdot d_x^l \cdot f_ {seg}\left(x_{l}\right)+f_{seg}(x)
\label{eq:fs(p)}
\end{equation}
\end{footnotesize}%
where $f_{seg}(x_{l})$ represents the segmentation features corresponding to $l$, $W_{M}(x)$ represents the multi-scale adaptive weights of different $S$ and $S$ denotes set of scales(e.g. 3×3, 9×9, 15×15). For these three different affinity field sizes, $N$ is set as 24 to include more affinity relative location. $W_{M}(x)$ are weights learned by one convolution layer after the affinity decoder.
We utilize $W_{M}$ together with the multi-scale affinity fields, to capture the semantic relationships with cross-scale neighbors. 
According to the exemplar results of $W_{M}$ in Fig.~\ref{weight}, the 3×3 weight map focuses more on the boundary and detailed information as these regions have stronger affinity associations with their neighboring regions. As the scale of the affinity field increases, more large-scale structural information would be taken into consideration. As a result, the 9×9 weight map focuses on more vessel main branches and the 15×15 weight map contains more global information for each pixel. Given vessels with various thicknesses, jointly utilizing the multi-scale affinity relationships as context and geometric guidance surely is helpful.

It is noteworthy that SMAFS relies on the intrinsic semantic label relationship instead of the absolute pixel feature intensity, which is more robust to visual appearance variations across images.% 

\subsection{Unsupervised Affinity Feature Strengthening}
Though the small objects have been claimed to own low responses at lower resolution features\cite{gao2019ssap}, enforcing unsupervised affinity feature strengthening (UAFS) may also be helpful.
As the affinity field ground truth corresponding to the down-sampled features is unavailable, UAFS is deployed to predict affinity relationships among pixels in different layers for feature enhancement in an implicit way. In this way, UAFS enhances segmentation features in different resolution features which would further establish affinity relationships among neighboring pixels for feature enhancement of deeper layers.

Compared to SMAFS, UAFS builds similar category location affinity field $d_x^l$ only in the 3×3 scale to fuse the segmentation features $f_{seg}(x)$ without affinity field supervision as in Eq.~\eqref{eq:AFS}.

\begin{footnotesize}
\begin{equation}
% f_{s}(x)=\sum_{l \in \mathcal{N}(x)} y_{A_{sl}}(x) \cdot f_{\mathrm{seg}}\left(x_{l}\right)+f_{\mathrm{seg}}(x)
f_{\mathrm{s}}(x)=\sum_{l \in N(x)}  d_x^l \cdot f_{seg}\left(x_{l}\right)+f_{seg}(x)
\label{eq:AFS}
\end{equation}
\end{footnotesize}%

According to Eq.~\eqref{eq:AFS}, UAFS utilizes neighboring pixel features and directly predicts a single-scale affinity field $y_x^l$ without supervision to strengthen segmentation features. For the calculation of $d_x^l$ in UAFS, we utilize $y_x^l$ to construct $d_x^l$ in the same way as SMAFS in Eqs.~\eqref{eq:Amean} and~\eqref{eq:Hard}. This establishes semantic relationships of neighboring pixels for encoding richer context and geometric information. In this way, UAFS can combine both semantic features and affinity information in deeper layers of the network to better encode more geometric information and establish richer relationships among features that are more robust to contrast change. 
\begin{figure*}[tbp]
	\centering{\includegraphics[width=1 \textwidth]{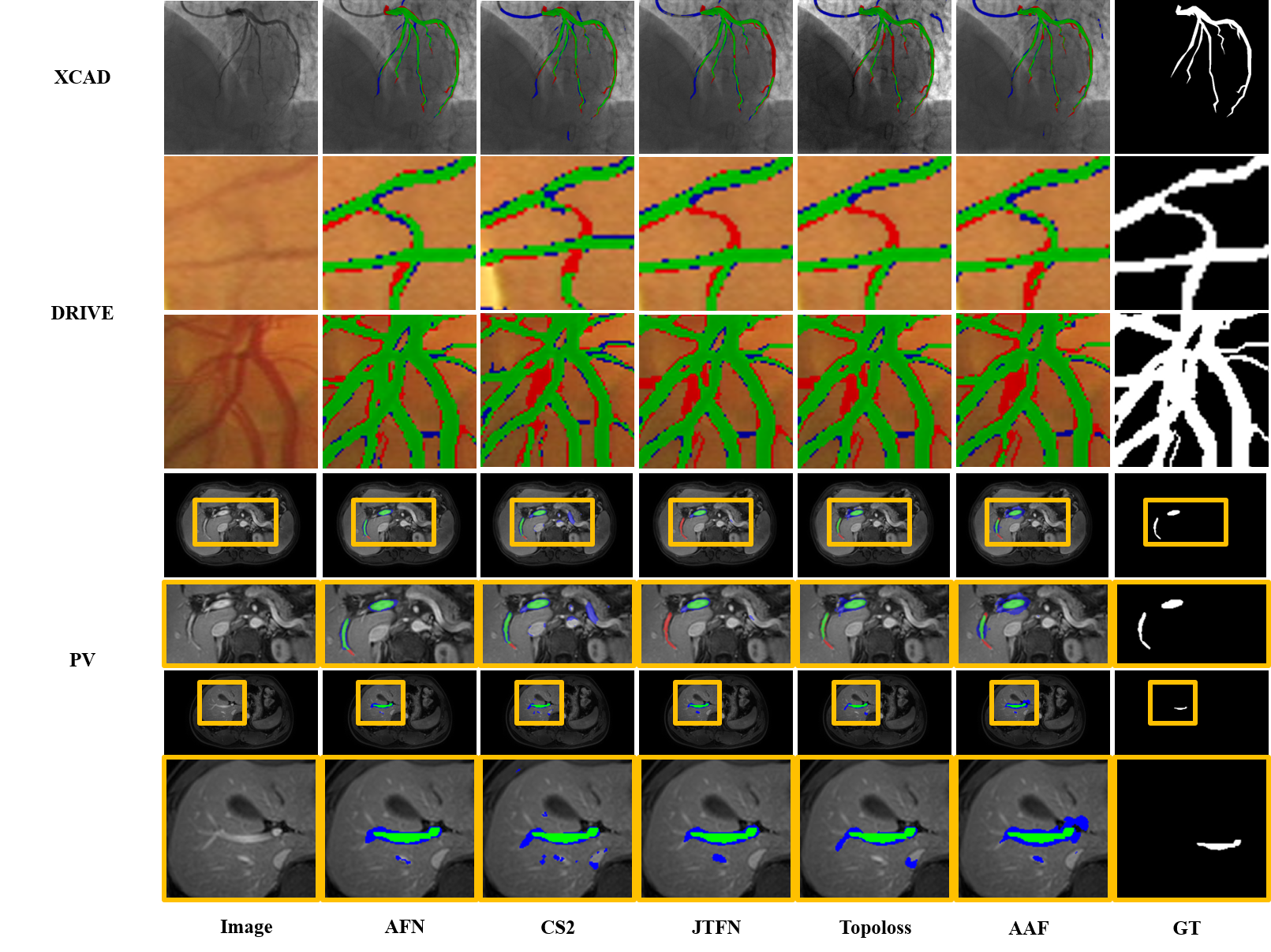}}
	\caption{Exemplar segmentation results of AFN and alternatives. Green pixels: TPs; Red pixels: FNs; Blue pixels: FPs. The yellow box indicates the local zoom area.} 
	\label{indataset_fig}
\end{figure*}

\section{Experiments}
\label{sec:experiment}
\subsection{Datasets}
\label{subsec:datasets}
\textbf{XCAD}: An X-ray angiography coronary artery disease (XCAD) dataset which include coronary angiography images obtained during stent placement using a General Electric Innova IGS 520 system\cite{ma2021self}. Each image has a resolution of 512$\times$512 pixels with one channel. This dataset contains 126 independent coronary angiograms with vessel segmentation maps annotated by experienced radiologists, which are divided into 84 training images and 42 test images with vessel segmentation maps annotated by experienced radiologists. Low-power X-ray and contrast agent doses are used during X-ray coronary angiography, leading to noisy and low-contrast coronary angiograms\cite{ma2021self}. In our experiment, we utilize artificial contrast perturbations to evaluate the robustness of our method against contrast variations.

\textbf{DRIVE}: 
The DRIVE dataset\cite{maninis2016deep} is a retinal vessel segmentation dataset consisting of 40 color retinal images with the same resolution as 565×584 pixels. Following\cite{cheng2021joint}, we set 20 images for training and 20 images for testing.

\textbf{Portal Vein}: Portal vein vessel dataset (PV) is an in-house dataset containing 32 patient cases. All these cases were captured by a 3.0T MRI scanner (Philips, Netherlands standardized protocols, using the following parameters: voxel size, 1.75$\times$1.75$\times$3.5; Reconstruction matrix, 352; flip angle, 10 deg). These MRI images have different resolutions varying from 352$\times$352 to 448$\times$448 pixels. For a fair evaluation, we randomly divided the dataset into 24 cases for training and 8 cases for testing. The whole dataset is annotated by experienced radiologists.
% (Philips, Netherlands standardized protocols, using the following parameters: voxel size, 1.75$\times$1.75$\times$3.5; Reconstruction matrix, 352; flip angle, 10 deg).

\textbf{DSA vessel}: This dataset is an in-house dataset consisting of 20 slices with the same resolution of 512$\times$512 pixels from different phases in digital subtraction angiography cerebrovascular vessel acquired with  a biplane angiography suite (Artis zee, Siemens, Forchheim). The annotation process of this dataset is time-consuming and extremely laborious. Thus, this dataset is only annotated with the clearest vessel by experienced radiologists.
Due to the contrast variations of DSA images from different frames, it can be used to evaluate the robustness of different methods against contrast change across images.
% Due to the high labeling cost, this dataset is used to testify the generalization performance.
%These DSA images were performed in a biplane angiography suite (Artis zee, Siemens, Forchheim).

The in-house PV and DSA datasets have been reviewed and approved by the institutional review boards of the Medical Ethics Committee of the Union Hospital, Tongji Medical College, Huazhong University of Science and Technology. All the centers registered and approved the studies labeled as Project Number 2022 (0311).
\subsection{Implementation Details}
\label{subsec:details}
\textbf{Evaluation protocol}: In order to make a fair comparison as \cite{cheng2021joint} to evaluate both the topology-level and pixel-level performance for vessel segmentation, we choose two kinds of metrics including: F1 score, Precision and Recall for pixel-level evaluation, and the Correctness, Completeness and Quality metrics in\cite{wiedemann1998empirical} for topology-level evaluation which measures the similarity between the predicted skeletons and the ground truth within a threshold (more detail shown in Appendix.~\ref{subsec:topometric}). Here, Correctness and Completeness can be regarded as the Precision and Recall metrics for skeleton similarity while Quality is a combination of Completeness and Correctness defined as 

\begin{footnotesize}
\begin{equation}
\mathrm{Quality} = \frac{\mathrm{Complete.} \cdot \mathrm{Correct.}}{\mathrm{Complete.}-\mathrm{Complete.} \cdot \mathrm{Correct.}+\mathrm{Correct.}}
\label{eq:completeness}
\end{equation}
\end{footnotesize}%
In our experiments, the threshold is set to 1 for the DRIVE dataset and 2 for other datasets.

\textbf{Training details}: Images are first augmented via horizontal flipping, random brightness and contrast range from 1.0 to 2.1, random saturation range from 0.5 to 1.5, and random rotation with 90°, 180°, and 270°, and then cropped to 256×256 pixels for training. All the networks were trained using an Adam optimizer with an initial learning rate of $10^{-3}$ with a weight decay of $5\times10^{-4}$ and a batch size of 2 for 2000 epochs. The hyper-parameter $\lambda_{b}$ in Eq.~\eqref{eq:TotalLoss} is set as 5. The affinity field scale lists are [3,5,7] for DRIVE and [3,9,15] for XCAD. The total amount parameter of AFN is 38.51M. The computational requirements FLOPS of AFN is 272.726G.

\subsection{Comparison with State-of-the-art}
\label{subsec:indataset}
We compare our AFN with various state-of-the-art methods on three-vessel datasets, including XCAD, DRIVE and PV.

For comprehensive evaluation, the most representative state-of-the-art pixel-level (CS$^{2}$Net\cite{mou2021cs2}), topology-level (Topoloss\cite{hu2019topology}), hybrid-level (JTFN\cite{cheng2021joint}), and affinity field feature learning (AAF\cite{ke2018adaptive}) based methods are selected for comparison. Table~\ref{indatasettab-XCAD} shows the performance of AFN and alternatives on the XCAD, DRIVE, and PV datasets respectively. Across different datasets, AFN consistently outperforms these state-of-the-art methods in terms of most metrics, leading to better performance in both pixel and topology levels. Specifically, AFN improves the Quality score by up to $7.03\%$ and the F1 score by up to $3.72\%$ compared to the second-best approach on the XCAD dataset. Exemplar qualitative results are given in Fig.~\ref{indataset_fig}. Compared to other methods, AFN achieves better vessel connectivity with fewer false positives, resulting in better pixel-level and topology-level segmentation performance. AFN can even segment the real vessels (denoted by blue pixels), which are of very low contrast and ignored by the manual annotation in Fig.~\ref{weight} and Fig.~\ref{indataset_fig}. These ”false positives” in turn would degrade the segmentation performance according to the classical pixel-level metrics, such as Precision.

\begin{table}[thbp]
\caption{Quantitative in-dataset evaluation of AFN compared with different methods on XCAD, DRIVE and Portal Vein datasets.}\label{indatasettab-XCAD}
\centering
\setlength\tabcolsep{3pt} 
\begin{tabular}{c|c|cccccc}
%\toprule[1pt]
\hline
Datasets & Methods             & Precision    & Recall      & F1      & Correct.      & Complete.     &Quality\\ \hline
\multirow{5}{*}{XCAD} & AAF  & 73.96& 77.91 & 75.51   & 71.30   & 79.22 &59.76\\
~ & Topoloss  & 75.08& 81.15 & 77.70  & 70.12   & 83.42 &61.07\\  
~ & CS$^{2}$Net& 74.95 & 78.79  & 76.42   & 74.25 & 81.22 &63.13\\ 
~ & JTFN & 78.11& 79.55 & 78.45   & 76.87   & 82.78 &65.86\\
~ & AFN  & \color{red}{\textbf{81.05}} & \color{red}{\textbf{83.95}} & \color{red}{\textbf{82.17}}   & \color{red}{\textbf{84.44}}   & \color{red}{\textbf{84.53}} &\color{red}{\textbf{72.89}}\\
\hline
\multirow{5}{*}{DRIVE} & AAF  & 81.49& 72.10 & 76.24   & 49.07   & 35.84 &26.10\\
~ & Topoloss  & 82.94& 80.29 & 81.36  & 55.67   & 46.95 &34.22\\  
~ & CS$^{2}$Net& 78.59 & 81.95  & 80.01   & 54.78 & 46.11 &33.39\\ 
~ & JTFN & 82.71& 83.40 & 82.81   & 57.09   & 49.28 &36.02\\
~ & AFN  & \color{red}{\textbf{83.22}} & \color{red}{\textbf{83.50}} & \color{red}{\textbf{83.16}}   & \color{red}{\textbf{57.26}}   & \color{red}{\textbf{49.40}} &\color{red}{\textbf{36.10}}\\
\hline
\multirow{5}{*}{PV} & AAF  & \color{red}{\textbf{84.86}}& 75.65 & 78.04   & 82.75   & 84.05 &70.42\\
~ & Topoloss  & 76.29& 76.67 & 73.67  & 70.87   & 81.84 &60.68\\  
~ & CS$^{2}$Net& 80.36 & 78.50  & 77.93  & 75.82 & \color{blue}{\textbf{87.96}} &68.49\\ 
~ & JTFN & 81.90& 82.02 & 80.63   & 79.84   & \color{red}{\textbf{91.43}} &70.76\\
~ & AFN  & \color{blue}{\textbf{83.15}} & \color{red}{\textbf{83.56}} & \color{red}{\textbf{82.07}}   & \color{red}{\textbf{85.58}}   &\textbf{86.71} &\color{red}{\textbf{73.42}}\\
\hline
%\bottomrule[1pt]
\end{tabular}
\end{table}

\begin{table}[thbp]
\caption{Ablations of MAFS and AFS for AFN on XCAD dataset.}\label{ablation_study}
\centering
\setlength\tabcolsep{3pt} 
\begin{tabular}{c|cccccc}
%\toprule[1pt]
\hline
Methods             & Precision    & Recall      & F1      & Correct.      & Complete.     &Quality\\ \hline
Base  & \color{red}{\textbf{81.30}}& 80.44 & 80.49   & 82.30   & 82.19 &69.34\\
SSAFS  & 80.91& 81.22 & 80.74  & 80.88   & 82.23 &68.47\\  
SSAFS+UAFS& 80.33 & 82.58  & 81.05   & 82.23 & 84.47 &71.08\\ 
SMAFS & 80.34& 81.35 & 80.50   & 84.38   & 81.19 &70.09\\\cline{1-7}
AFN  & \color{blue}{\textbf{81.05}} & \color{red}{\textbf{83.95}} & \color{red}{\textbf{82.17}}   & \color{red}{\textbf{84.44}}   & \color{red}{\textbf{84.53}} &\color{red}{\textbf{72.89}}\\
\hline
%\bottomrule[1pt]
\end{tabular}
\end{table}
\subsection{Ablation Study}
\label{subsec:ablation}
To validate the effectiveness of each component in AFN, we conduct ablation studies on the XCAD dataset, including \textbf{Base}: The baseline model of AFN by removing supervised multi-scale affinity feature strengthening (SMAFS) and unsupervised affinity feature strengthening (UAFS).

\textbf{Supervised Single-scale Affinity Feature Strengthening (\textbf{SSAFS})}: The model in which supervised single-scale affinity feature strengthening (i.e., 5×5 scale) is introduced to the Base. According to Table~\ref{ablation_study}, SSAFS achieves slight improvements on Recall, Complete and F1, indicating the limitation of a single-scale affinity field for feature strengthening.

\textbf{Supervised Multi-scale Adaptive-affinity Feature Strengthening (\textbf{SMAFS})}: A model in which we integrate the SMAFS module with three scales into the Base. Comparing the results of SMAFS and SSAFS in Table~\ref{ablation_study}, we can see noticeable improvements in terms of Recall, Correct and Quality. Especially for Quality, SMAFS is $0.75\%$ and $1.62\%$ higher than those of Base and SSAFS respectively. It shows that the value of multi-scale affinity feature strengthening establishes more geometric relationships among pixels. However, as feature strengthening is only implemented on the highest resolution features, it can hardly improve the pixel-level segmentation performance plenty.

\textbf{Unsupervised Affinity Feature Strengthening (UAFS)}: A model in which we integrate both SSAFS and UAFS into the Base. As shown in Table~\ref{ablation_study}, SSAFS+UAFS performs $0.31\%$ higher in F1 and $2.61\%$ higher in Quality compared to SSAFS. The above results prove that UAFS can consistently strengthen the segmentation features by implicit affinity feature strengthening. 

\textbf{Including both SMAFS and UAFS (AFN)}: According to Table~\ref{ablation_study}, jointly using SMAFS and UAFS achieves the best overall performance, outperforming Base by up to $1.68\%$ in F1 and $3.55\%$ in Quality. As discussed above, ``over-segmentation" (i.e. detecting more true but unannotated vessels) would be counter-productive according to classical pixel-wise evaluation metrics like Precision. It explains why the Precision score of AFN is slightly lower while the overall performance is much better. Compared to different component combinations, combining SMAFS and UAFS can perform feature strengthening most effectively and improve both the pixel-level and the topology-level performance simultaneously.

\section{Discussion}
\subsection{Evaluation of AFN with varying sizes}
\label{subsec:thinthick}
\begin{figure*}[tbp]
	\centering{\includegraphics[width= \textwidth]{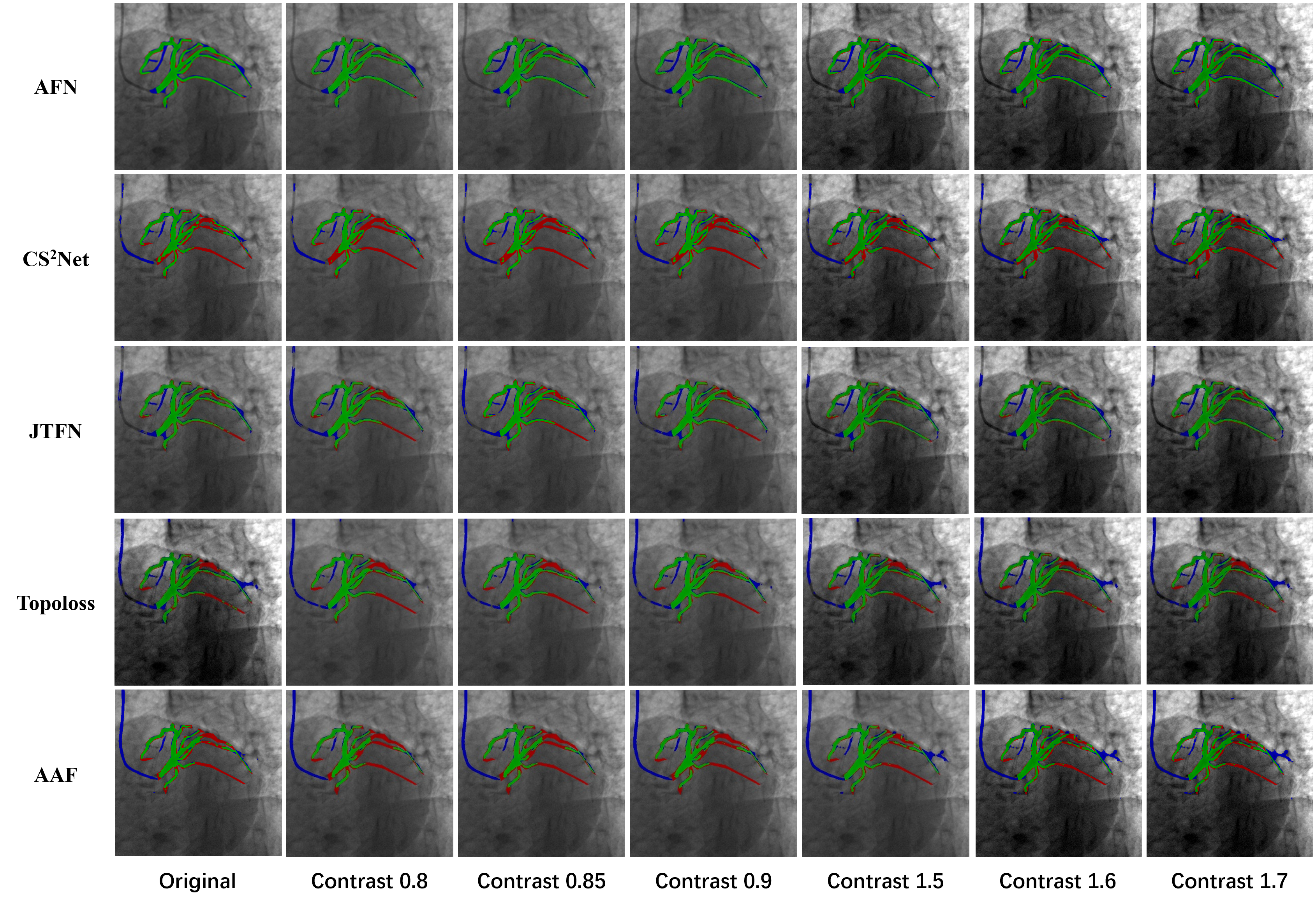}}
	\caption{Exemplar segmentation results under various contrast in XCAD. Green pixels: TPs; Red pixels: FNs; Blue pixels: FPs.}
	\label{xcadcontrast_fig}
\end{figure*}
\begin{figure*}[htbp]
	\centering{\includegraphics[width= \textwidth]{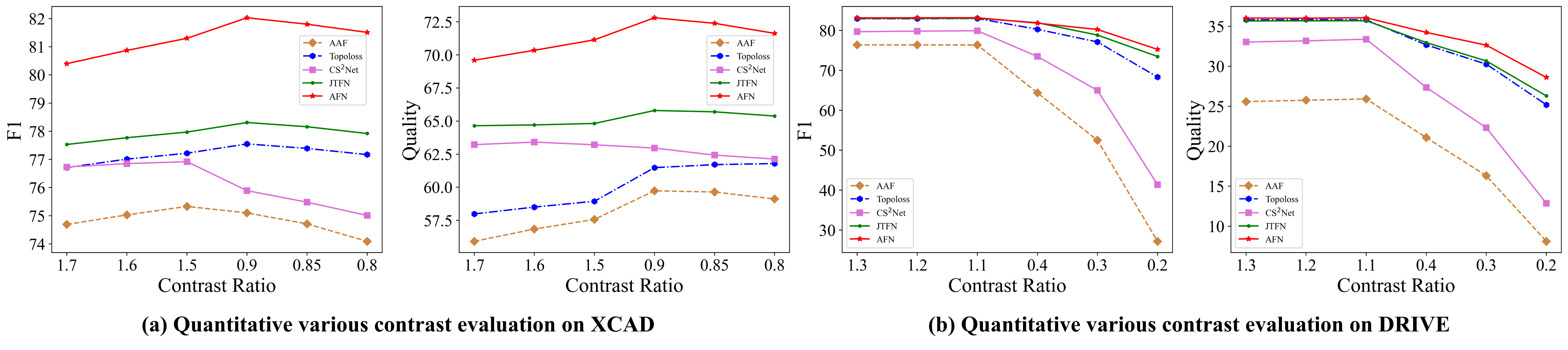}}
	\caption{Quantitative various contrast evaluation compared with state-of-the-art methods}
	\label{xcadcontrastcurve_fig}
\end{figure*}
Accurately segmenting vessels of different sizes, in particular for thin vessels, are quite challenging. To this end, we set the vessels thinner than seven pixels as thin vessels and the rest vessels as thick vessels for the XCAD dataset. %Similarly, for DRIVE, the threshold is set as 3 pixels. 
Then, each thin vessel is assigned with a 5-pixel searching range and pixels in the given segmentation map located within the range are counted for pixel-to-pixel matching, while each thick vessel is assigned with a 10-pixel searching range for pixel-to-pixel matching as\cite{yan2018joint}.

Quantitative segmentation results of both thick and thin vessels on the two datasets are summarized in Table~\ref{thinthick-XCAD}. Here, only Correctness, Completeness, Quality and F1 are selected as evaluation metrics to emphasize more on topological completeness (F1 is a balance metric for pixel-level and Quality is a balance metric for topology-level ). Compared to different methods, AFN achieves the best overall segmentation performance, especially in F1, Correctness and Quality. 
%Though AFN performs slightly worse on Correct of DRIVE(58.01 vs 58.31) for thin vessels compared with TopoNet (specially designed for topology-level performance), AFN strikes a better balance between topology-preservation and pixel-level segmentation. 

\textbf{Robustness to Various Sizes} 
Our AFN adopts different multi-scale adaptive weights for the SMAFS module to capture the semantic relationships with cross-scale neighbors. Specifically, the smaller weight map focuses more on the boundary and detailed information as these regions have stronger affinity associations with their neighboring regions. As the scale of the affinity field increases, more large-scale structures are taken into consideration. Given vessels with various thicknesses, jointly utilizing the multi-scale affinity relationships as context and geometric guidance is helpful. Thus, our AFN brings stable performance improvements for various sizes of vessels.
\begin{table}[thbp]
\caption{Quantitative results of different methods on thick and thin vessel segmentation on the XCAD dataset.}\label{thinthick-XCAD}
\centering
\setlength\tabcolsep{3pt} 
\begin{tabular}{c|c|cccc}
%\toprule[1pt]
\hline
Thickness &Methods              & F1      & Correct.      & Complete.     &Quality\\ \hline
\multirow{5}*{Thin} & AAF  & 81.23   & 94.00   & 78.23 &74.49\\
~&Topoloss   & 82.85  & 94.07   & 82.83 &78.69\\  
~&CS$^{2}$Net & 80.91   & 92.47 & 80.60 &75.60\\ 
~&JTFN  & 81.83   & 93.88   & 82.23 &78.01\\
~&AFN  & \color{red}{\textbf{83.67}}   & \color{red}{\textbf{94.64}}   & \color{red}{\textbf{83.51}}
&\color{red}{\textbf{79.74}}\\
\cline{1-6}
\multirow{5}{*}{Thick} &AAF   & 88.76   & 88.94   & 84.32 &76.94\\
~&Topoloss   & 90.14  & 89.03   & \color{red}{\textbf{86.44}} &78.79\\  
~&CS$^{2}$Net  & 88.32   & 86.53 & 82.72 &74.01\\ 
~&JTFN  & 88.52   & 90.14   & 84.77 &78.26\\
~&AFN  & \color{red}{\textbf{90.62}}   & \color{red}{\textbf{92.04}}   & \color{blue}{\textbf{84.98}}
&\color{red}{\textbf{79.61}}\\
\hline
%\bottomrule[1pt]
\end{tabular}
\end{table}%

\subsection{Evaluation of AFN with contrast changes}
\label{subsec:contrast-insensitive}
In clinical scenarios, various image contrast often degrades the performance of deep learning-based methods. In this section, we evaluate the robustness of the proposed AFN against contrast variations. We follow the approach in \cite{hendrycks2018benchmarking} to edit image contrast on XCAD and DRIVE defined as

\begin{footnotesize}
\begin{equation}
I_{contrast(x)}=I_{average}+\left(I(x)-I_{average}\right) \cdot \text{ Contrast  Ratio}
\label{eq:contrast}
\end{equation}
\end{footnotesize}%
where $I_{contrast}(x)$ and $I(x)$  represent the adjusted intensity and the original intensity of pixel $x$, $I_{average}$ represents the average intensity of the whole image. In terms of the contrast ratio, it is sampled from [1.7,1.6,1.5,0.9,0.85,0.8] for XCAD and [1.3,1.2,1.1,0.4,0.3,0.2] for DRIVE respectively. Then, the images with varying contrast are used to evaluate different methods as depicted in Fig.~\ref{xcadcontrastcurve_fig} (more segmentation result in DRIVE shown in Appendix.~\ref{subsec:driveintesities}). Compared with the state-of-the-art methods, AFN achieves the best results in F1 and Quality, indicating better pixel-level and topology-level segmentation performance. In addition, across different levels of contrast change, AFN is also with the least performance degradation. Compared to the most competitive method JTFN, AFN achieves significant improvements of $3.59\%$ in F1 and $7.01\%$ in Quality on XCAD and $1.79\%$ in F1 and $2.31\%$ in Quality for DRIVE. Exemplar qualitative results on XCAD are shown in Fig.~\ref{xcadcontrast_fig}. AFN stably produces a more complete vascular structure under different contrasts, demonstrating the robustness of AFN when applied to vessel segmentation in real clinical scenarios.

\begin{figure}[tbp]
	\centering{\includegraphics[width=0.5 \textwidth]{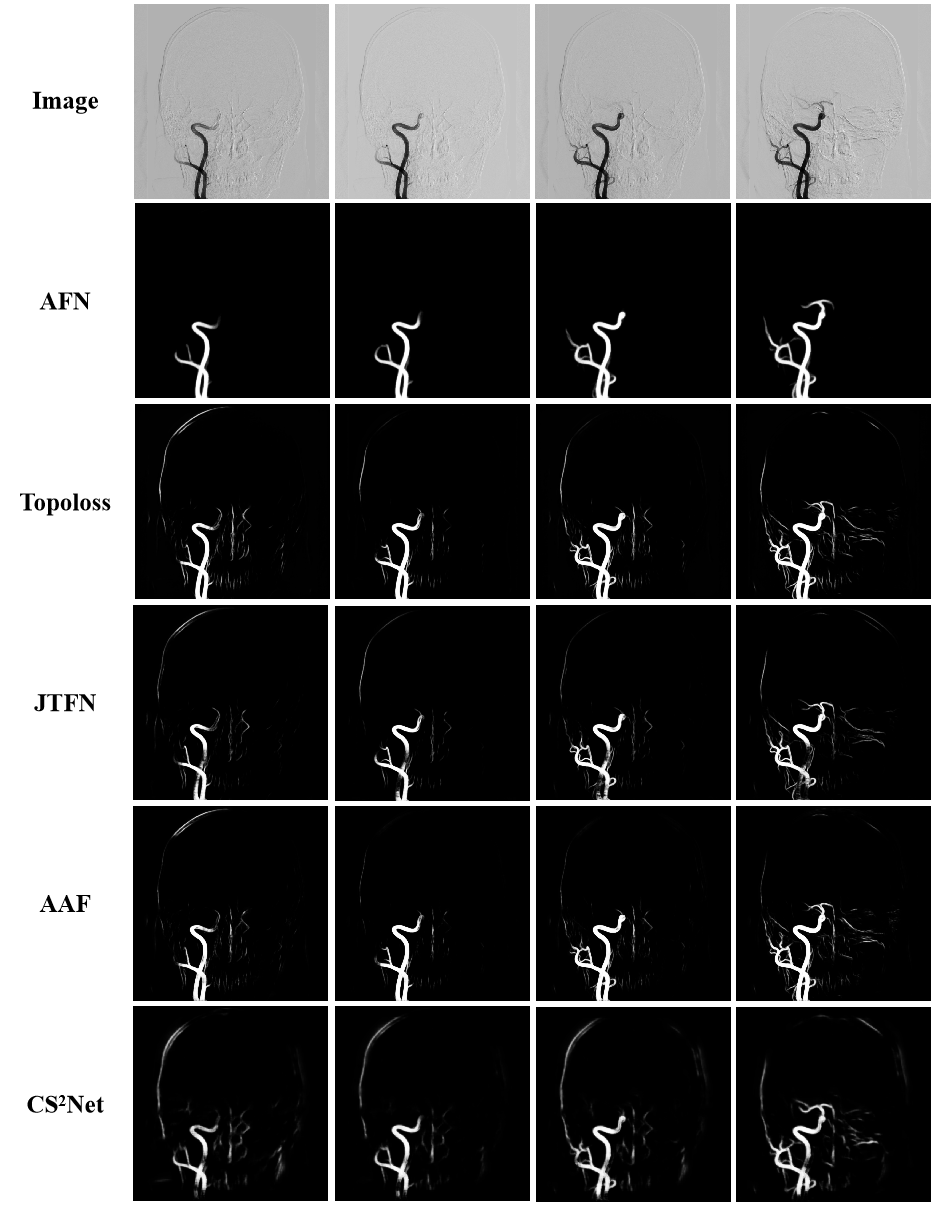}}
	\caption{Exemplar segmentation results for various contrast on DSA.}
	\label{dsacontrast_fig}
\end{figure}

To validate this, we conduct an additional evaluation of the vessel segmentation of DSA images, as there exists natural contrast change between DSA images from different frames. We use the models trained on XCAD directly to the test DSA images as shown in Fig.~\ref{dsacontrast_fig}. Despite of DSA contrast agent changing, AFN achieves the most robust results, not only for blood vessels but also for some disturbances (such as skull bones). 

Though image contrast can change across different situations, the semantic and geometric structure relationships among pixels should be consistent and this information in turn makes AFN more robust against absolute pixel intensity change. It explains why AFN is more stable than other methods in terms of contrast change.

\textbf{Robustness to Contrast Changes}
The proposed SMAF and UAFS modules establish semantic and geometric structure relationships among different scale neighboring pixels for feature enhancement of different layers. As such relationships remain consistent in different contrast scenarios, AFN can well achieve good robustness to contrast changes. Furthermore, AFN has better feature representation capability by enhancing features of each pixel using an aggregation of its neighboring pixel features.

\subsection{Evaluation of generalizability}
\label{subsec:generalizability}
To further demonstrate the generalizability of the proposed AFN, we apply the models trained on XCAD to the DSA dataset for testing. Though DSA also consists of angiography images, there exist large contextual/appearance variations compared to XCAD. As stated in Table~\ref{generalize_tab}, compared to other methods, AFN achieves the best overall segmentation performance, resulting in significant improvements in terms of F1 ranging from $3.63\%$ to $16.71\%$. For topology-level performance, AFN outperforms other methods by a large margin (up to $11.64\%$ in Quality). 

\begin{figure}[tbp]
\centering{\includegraphics[width=0.5 \textwidth]{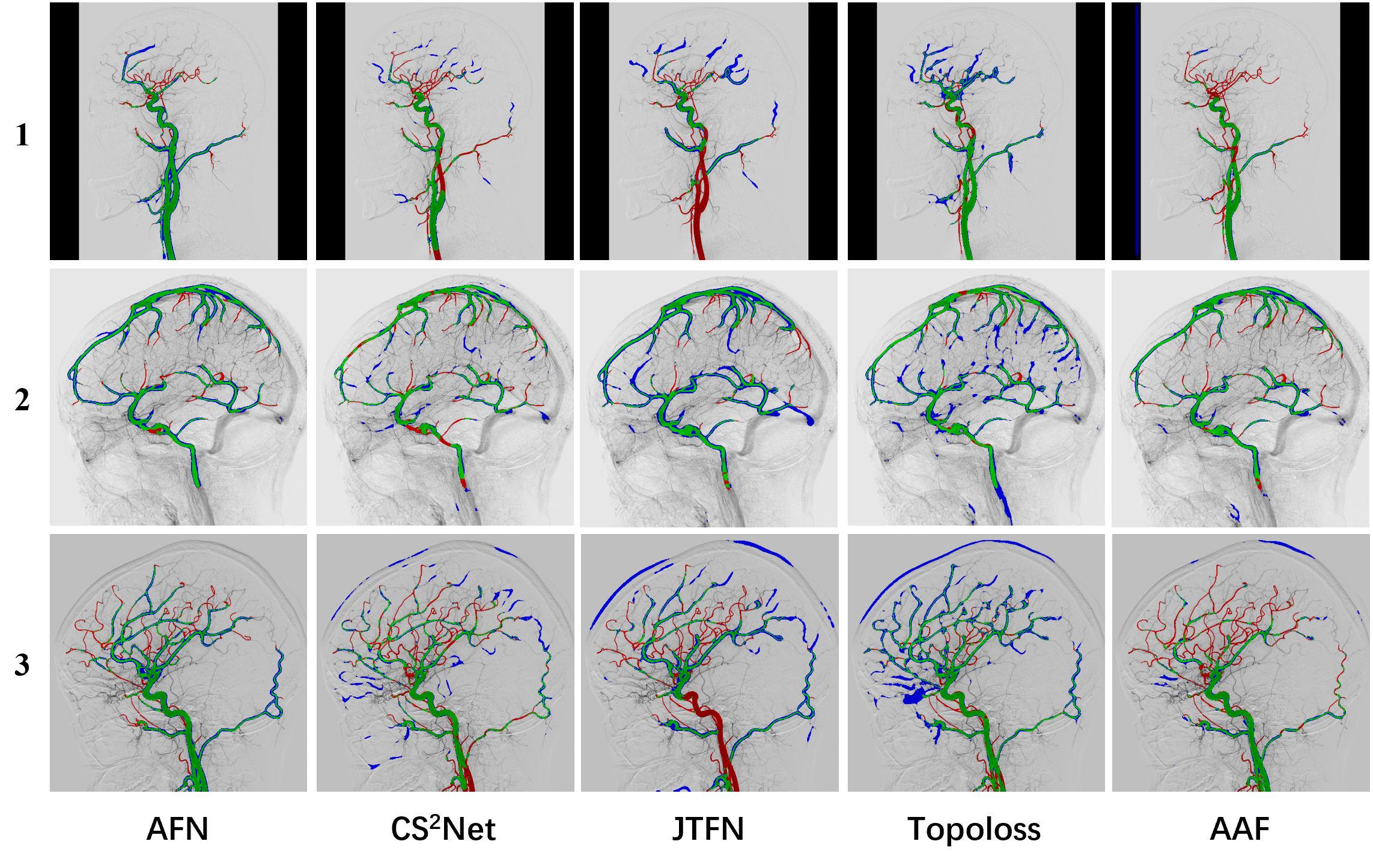}}
\caption{Exemplar segmentation results on DSA using the model trained on XCAD. Green pixels: TPs; Red pixels: FNs; Blue pixels: FPs.}
\label{generalize_fig}
\end{figure}

\textbf{Robustness to Domain Changes} According to the qualitative results of the arterial phase 1, venous phase 2 and capillary phase 3 shown in Fig.~\ref{generalize_fig}, AFN is proven to be capable  of capturing true vessel structures across different domains. It is because AFN can learn more relevant structure and semantic relations for feature strengthening, which helps better distinguish vessels and alleviate vessel-like interferences (such as skull bone structures). The above results demonstrate the generalization ability of the proposed AFN for better deployment.

\begin{table}[thbp]
\caption{Quantitative generalization evaluation of AFN compared with different methods on XCAD->DSA.}\label{generalize_tab}
\centering
\setlength\tabcolsep{4pt} 
\begin{tabular}{c|cccccc}
%\toprule[1pt]
\hline
Methods             & Precision    & Recall      & F1      & Correct.      & Complete.     &Quality\\ \hline

AAF  & 68.38& 69.16 & 68.09   & 74.12   & 49.51 &41.57\\
Topoloss  & 68.98& 72.84 & 69.53  & 79.75   & \color{red}{\textbf{60.27}} &50.88\\  
CS$^{2}$Net& \color{red}{\textbf{73.92}} & 67.69  & 70.02   & 82.02 & 55.98 &49.28\\ 
JTFN & 55.39& 60.37 & 56.94   & 76.90   & 48.45 &41.83\\
AFN  & \color{blue}{\textbf{70.13}} & \color{red}{\textbf{78.74}} & \color{red}{\textbf{73.65}}   & \color{red}{\textbf{91.63}}   & \color{blue}{\textbf{56.42}} &\color{red}{\textbf{53.21}}\\
\hline
%\bottomrule[1pt]
\end{tabular}
\end{table}

\par
\subsection{Limitation and Future Work}
\label{subsec:Limitation}
% \textbf{Limitation and Future Work}
In this work, AFN was designed in a 2D manner without exploring cross-slice information for 3D vessel segmentation. 
Through extensive comparison experiments, the affinity field is proven effective for feature enhancement of each pixel using an aggregation of its neighboring pixel features. 
In 3D vessel segmentation, the affinity field can be directly applied by extending the neighborhood of each pixel from 2D to 3D. In this way, the affinity field could work as a context and geometric guidance for 3D blood vessel feature enhancement. In the future, we will extend AFN to more vessel segmentation applications.

\section{Conclusion}
\label{sec:Conclusion}
We propose an Affinity Feature Strengthening Network (AFN) for vessel segmentation. The cores of our AFN are one SMAFS and three UAFS modules which utilize affinity fields to encode semantic relationships as the geometric constraint for the segmentation feature enhancement. Extensive experimental results on several datasets demonstrate that our AFN consistently outperforms state-of-the-art methods in terms of higher pixel-level accuracy, better topological completeness and robustness to contrast changes. Besides, the cross-dataset evaluation also shows the impressive generalizability performance of AFN.

\bibliographystyle{IEEEtran}
\bibliography{main}
\begin{appendix}
\label{sec:Appendix}
\subsection{Definition of Topology Metrics}
\label{subsec:topometric}
\begin{figure}[tbp]
	\centering{\includegraphics[width=0.5 \textwidth]{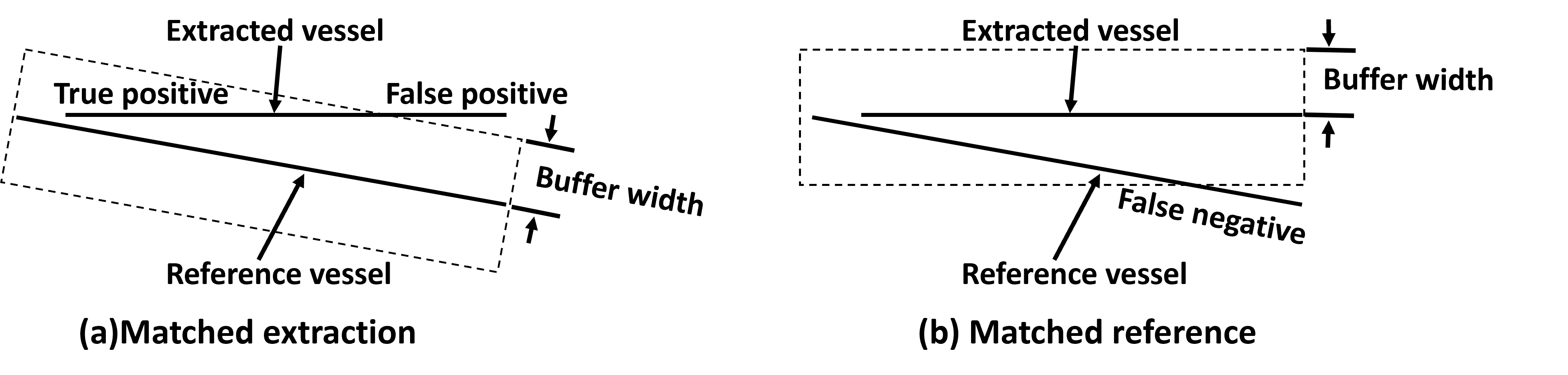}}
	\caption{Matching principle.}
	\label{metrics}
\end{figure}

As shown in Fig.~\ref{metrics}, we have defined the metrics of Completeness and Correctness. Fig.~\ref{metrics} illustrates the matching principle between the extracted vessel and the reference vessel based on which we calculate the metrics. The buffer of constant predefined width (buffer width) is constructed around the reference vessel data (see Fig.~\ref{metrics}a). Within this buffer any parts of the extracted data do not exceed a predefined threshold are considered as matched, and are labeled as true positive (TP) to signify that the extraction algorithm has successfully identified vessel data. On the other hand, unmatched extracted data is labeled as false positive (FP), because the extracted vessel hypotheses are deemed to be incorrect. In the second step, the matching process is performed in reverse with buffer is now constructed around the extracted vessel data (see Fig.~\ref{metrics}b). Any parts of the reference data located within the buffer are considered as matched, while unmatched reference data are denoted as false negative (FN).
% \color{black}

\begin{footnotesize}
\begin{equation}
\mathrm{Completeness} = \frac{length \quad of \quad matched \quad reference}{length \quad of \quad reference} = \frac{\mathrm{TP}}{\mathrm{TP}+\mathrm{FN}}
\label{eq:completeness_app}
\end{equation}
\end{footnotesize}%

Completeness refers to the percentage of the reference vessel that can be accounted for by the extracted vessel, i.e. the percentage of the reference vessel which lies within the buffer around the extracted vessel.

\begin{footnotesize}
\begin{equation}
\mathrm{Correctness} = \frac{length \quad of \quad matched \quad extraction}{length \quad of \quad extraction} = \frac{\mathrm{TP}}{\mathrm{TP}+\mathrm{FP}}
\label{eq:correctness_app}
\end{equation}
\end{footnotesize}%

Correctness represents the percentage of the correctly extracted vessel, i.e., the percentage of the extracted vessel which lies within the buffer around the reference vessel.

\subsection{More Examples for DRIVE dataset with different intensities}
\label{subsec:driveintesities}
\begin{figure}[tbp]
	\centering{\includegraphics[width=0.5 \textwidth]{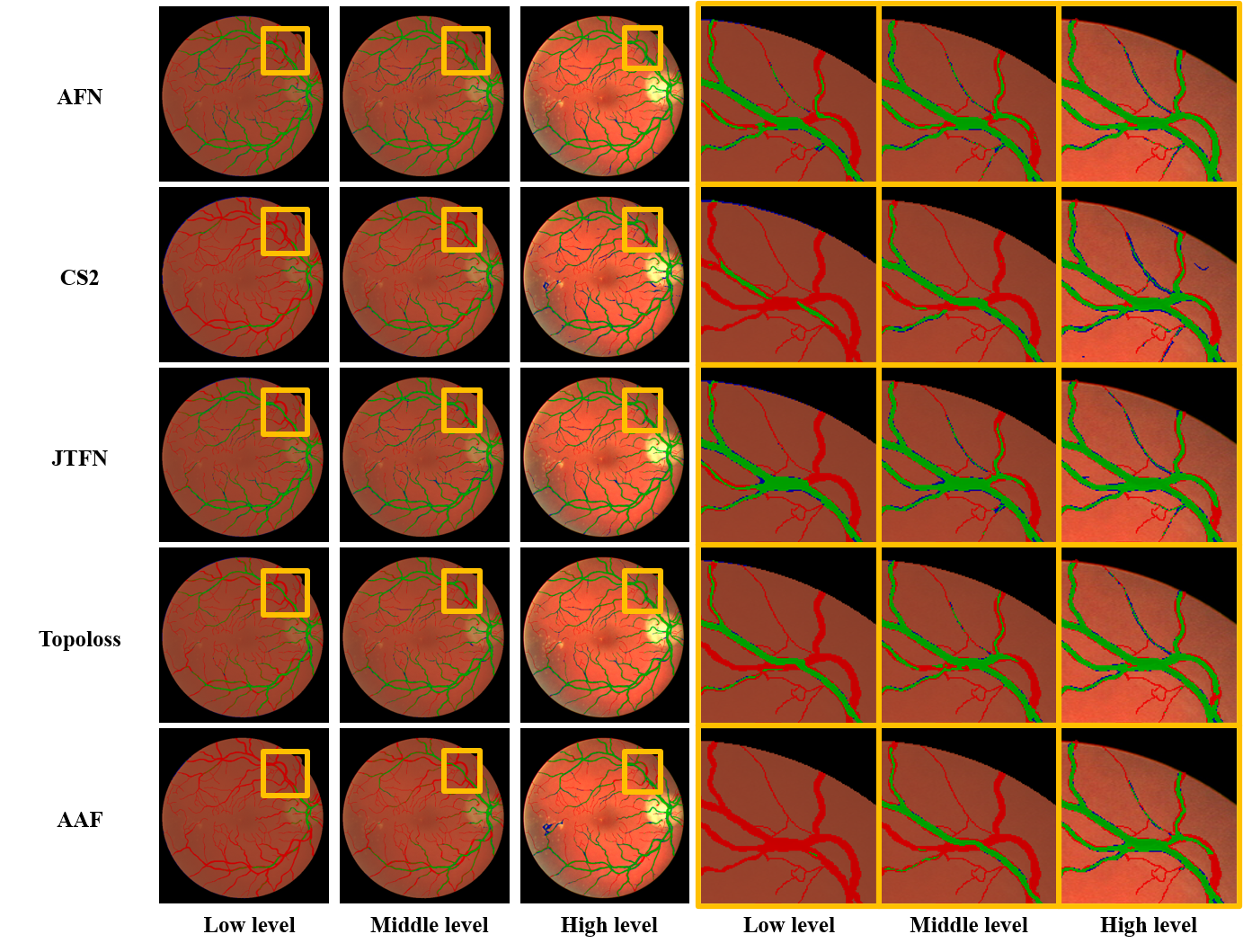}}
	\caption{Exemplar segmentation results of AFN and alternatives on DRIVE. Green pixels: TPs; Red pixels: FNs; Blue pixels: FPs. The yellow box indicates the local zoom area.}
	\label{contrast_fig}
\end{figure}

As shown in Fig.~\ref{contrast_fig}, more example on DRIVE dataset with 3 different intensities (Low level: contrast 0.2, Middle level: original image, High level: contrast 1.1 on DRIVE) are shown and zoomed the segmentation maps for better visualization performance in Fig.~\ref{contrast_fig}.
\end{appendix}
\end{document}